# Learning Belief Networks in Domains with Recursively Embedded Pseudo Independent Submodels

**J. Hu    Y. Xiang**
Dept. of Computer Science, Univ. of Regina
Regina, Saskatchewan, Canada S4S 0A2
junhu, yxiang@cs.uregina.ca

## Abstract

A *pseudo independent* (PI) model is a probabilistic domain model (PDM) where proper subsets of a set of collectively dependent variables display marginal independence. PI models cannot be learned correctly by many algorithms that rely on a single link search. Earlier work on learning PI models has suggested a straightforward multi-link search algorithm. However, when a domain contains recursively embedded PI submodels, it may escape the detection of such an algorithm. In this paper, we propose an improved algorithm that ensures the learning of all embedded PI submodels whose sizes are upper bounded by a predetermined parameter. We show that this improved learning capability only increases the complexity slightly beyond that of the previous algorithm. The performance of the new algorithm is demonstrated through experiment.

## 1  INTRODUCTION

Learning belief networks has been researched actively by many as an alternative to elicitation in knowledge acquisition [3, 1, 4, 2]. A *pseudo-independent* (PI) model is a probabilistic domain model (PDM) where proper subsets of a set of collectively dependent variables display marginally independence (hence pseudo-independent) [8, 6]. Commonly used algorithms for learning belief networks rely on a single link lookahead search to identify local dependence among variables. These algorithms cannot learn correctly when the domain model unknown to us is a PI model [7]. If an incorrectly learned model is used for subsequent inference, it will cause decision mistakes. Worse yet, the mistakes will be made without even knowing. The pseudo-independent property of PI models requires *multi-link* lookahead search in order to detect the collective dependency [8]. As the computational complexity increases exponentially with the number of links to lookahead, a multi-link search must be performed cautiously. In order to manage the increased complexity, it is suggested [6] that the single link search should be performed first and then the number of links to lookahead should be increased one-by-one.

Several issues remain open. A straightforward multi-link lookahead search as suggested in [8] will perform a single link lookahead search, then a double link lookahead search, and then a triple link lookahead search, etc. It turns out that some PI models will escape such a multi-link search (to be detailed below). Therefore, Xiang [6] suggested to perform a single link lookahead search first, followed by a combination of double link lookahead and single link lookahead search, followed by a combination of triple, double and single link lookahead search, etc. However, it is unclear what is the most effective way to combine lookahead search of different number of links.

In this paper, we propose an algorithm for learning belief networks from PI domains. We focus on learning decomposable Markov networks [8], although the algorithm can be extended to learning Bayesian networks. We show that our algorithm will ensure correct learning of PI models that contain no embedded submodels beyond a predetermined size. The time complexity of the algorithm is analyzed.

We assume that readers are familiar with commonly used graph-theoretic terminologies such as connected graph, component of a graph, chordal graph, clique, I-map, Bayesian networks, Markov networks, etc.

The rest of the paper is organized as follows: In section 2, we briefly introduce PI models. In section 3, we present the algorithm. The property of the algorithm is analyzed in section 4. The complexity is analyzed in section 5. We present our experimental results in section 6.

## 2 BACKGROUND

To make this paper self-contained, we introduce the basic concepts on PI models briefly in this section. We will use freely the formal definitions in [6]. More detailed discussions and examples can be found in the above reference.

If each variable $X$ in a subset $A$ is marginally independent of $A \setminus \{X\}$, we shall say that variables in $A$ are *marginally independent*. A set $N$ of variables are *collectively dependent* if for each proper subset $A \subset N$, there exists no proper subset $C \subset N \setminus A$ such that $P(A|N \setminus A) = P(A|C)$. A set $N$ of variables are *generally dependent* if for any proper subset $A$, $P(A|N \setminus A) \neq P(A)$.

A *pseudo-independent* (PI) model is a probabilistic domain model (PDM) where proper subsets of a set of collectively dependent variables display marginal independence. PI models can be classified into three types. In a full PI model, every proper subset of variables are marginally independent.

**Definition 1 (Full PI model)** *A PDM over a set $N$ ($|N| \geq 3$) of variables is a* **full** *PI model if the following two conditions hold:*

(S1) *For each $X \in N$, variables in $N \setminus \{X\}$ are marginally independent.*

(S2) *Variables in $N$ are collectively dependent.*

In a partial PI model, not every proper subset of variables are marginally independent.

**Definition 2 (Partial PI model)** *A PDM over a set $N$ ($|N| \geq 3$) of variables is a* **partial** *PI model if the following three conditions hold:*

(S1') *There exists a partition $\{N_1, \ldots, N_k\}$ ($k \geq 2$) of $N$ such that variables in each subset $N_i$ are generally dependent, and for each $X \in N_i$ and each $Y \in N_j$ ($i \neq j$), $X$ and $Y$ are marginally independent.*

(S2) *Variables in $N$ are collectively dependent.*

In a PI model, it may be the case that not all variables in the domain are collectively dependent. An embedded PI submodel displays the same dependence pattern of the previous PI models but involves only a proper subset of domain variables.

**Definition 3 (Embedded PI submodel)**
*Let a PDM be over a set $N$ of generally dependent variables. A proper subset $N' \subset N$ ($|N'| \geq 3$) of variables forms an* **embedded** *PI submodel if the following two conditions hold:*

(S4) *$N'$ forms a partial PI model.*

(S5) *The partition $\{N_1, \ldots, N_k\}$ of $N'$ by S1' extends into $N$. That is, there is a partition $\{A_1, \ldots, A_k\}$ of $N$ such that $N_i \subseteq A_i$, ($i = 1, .., k$), and for each $X \in A_i$ and each $Y \in A_j$ ($i \neq j$), $X$ and $Y$ are marginally independent.*

In general, a PI model can contain one or more PI submodels, and this *embedding* can occur recursively for any finite number of times.

PDMs can often be concisely represented by a graph called an *I-map* [5] of the PDM. In this paper, we shall mainly use undirected I-maps. In particular, we focus on learning an I-map that is a *decomposable Markov network* (DMN). A DMN consists of a graphical structure and a probability distribution factorized according to the structure. The structure is a *chordal* graph whose nodes are labeled by domain variables.

Since variables in a PI submodel are collectively dependent, in a minimal I-map of the PDM, the variables in the submodel is completely connected. The marginal independence between subsets in the submodel is thus unrepresented. The undirected I-maps can be extended into *colored* I-maps [6]. The marginal independence between subsets are highlighted in a colored I-map by coloring the corresponding links.

**Definition 4** *An undirected graph $G$ is a* **colored I-map** *of a PDM $M$ over $N$ if (1) $G$ is a minimal I-map of $M$, and (2) for each PI submodel $m$, links between each pair of nodes from distinct marginally independent subsets in $m$ are colored. Other links are referred to as black.*

A partial PI model is shown in Table 1. The PDM has four variables, which are partitioned into three independent subsets. The PDM contains three embedded PI submodels over

$$N_1 = \{a, b, c\}, N_2 = \{d, a, c\}, N_3 = \{d, b, c\}.$$

Figure 1 shows the colored I-map of this model. The colored links are drawn as dotted. For example, from the distribution P(a, c, d), it is easy to verify that $N_2$ forms a partial PI submodel with the marginally independent partition $\{\{a\}, \{c,d\}\}$ (S4). This partition extends into a marginally independent partition $\{\{a\}, \{b, c, d\}\}$ (S5).

Table 1: A model with embedded PI submodels.

| $(d,a,b,c)$ | $P(.)$ | $(d,a,b,c)$ | $P(.)$ |
|---|---|---|---|
| (0,0,0,0) | 0.02 | (1,0,0,0) | 0.03 |
| (0,0,0,1) | 0.02 | (1,0,0,1) | 0.01 |
| (0,0,1,0) | 0.06 | (1,0,1,0) | 0.01 |
| (0,0,1,1) | 0 | (1,0,1,1) | 0.05 |
| (0,1,0,0) | 0.1 | (1,1,0,0) | 0.09 |
| (0,1,0,1) | 0.06 | (1,1,0,1) | 0.07 |
| (0,1,1,0) | 0.14 | (1,1,1,0) | 0.15 |
| (0,1,1,1) | 0.1 | (1,1,1,1) | 0.09 |

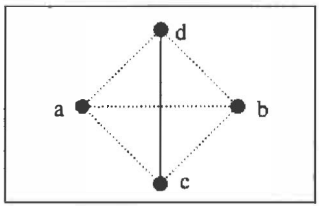

Figure 1: Colored I-map of the model in Table 1.

It has been shown [7] that common algorithms for learning belief networks cannot learn a PI model correctly because they rely on a single link lookahead search to identify local dependence among variables. For example, if these algorithms are used to learn the above model (assuming learning starts with an empty graph) only the link $(d,c)$ can be connected and the returned graph is not an I-map of the PDM.

## 3   THE LEARNING ALGORITHM

The pseudo independence property of PI models requires more sophisticated search procedures in learning. Suppose a PI submodel over $N' \subset N$ is partitioned into $k$ marginally independent subsets. If we lookahead by multiple links at each search step such that $N'$ is completely connected by a set of new links, and test $P(X|Y, N' \setminus X, Y) = P(X|N' \setminus X, Y)$, where $(X,Y)$ is one of the new links, we will get a negative answer. This prompts the completion of $N'$ in the learned graph. Based on this observation, a straightforward *multi-link search* is suggested in [8]. Such a search will perform a single link lookahead, followed by a double link lookahead, followed by a triple link lookahead, etc.

A multi-link search is more expensive than a single link search since $O(|N|^{2i})$ sets of links need to be tested before one set of links is adopted. Since the complexity increases exponentially with the number of links to lookahead, an multi-link search must be performed cautiously. Three strategies are proposed in [6] to manage the computational complexity: (1) performing single link search first, (2) increasing the number of links to search one-by-one, and (3) making learning inference-oriented.

Although the previous straightforward multi-link search can learn correctly many PI models, it was found that some PI submodels may still escape the learning algorithm. For example, if we apply such a search to the PI model in Table 1, the single link search will add the link $(d,c)$. The following double link search will first discover the PI submodel over $N_2$ and add links $(d,a)$ and $(a,c)$. It then discovers the PI submodel $N_3$ and add links $(d,b)$ and $(b,c)$. But the PI submodel over $N_1$ will never be learned by the double link lookahead or lookahead with higher number of links, since only a single link $(a,b)$ is unconnected. Consequently, the learning outcome will not be an I-map.

Realizing this deficiency of the straightforward multi-link search, an improved multi-link search algorithm was proposed in [6]. In addition to the incorporation of the above three strategies, the search is performed in the following manner: A single link lookahead is performed first, followed by a combination of double link lookahead and single link lookahead, followed by a combination of triple, double and single link lookahead, etc. We shall refer to such a systematic search that lookaheads by no more than $i > 1$ links as an *i-link search*. We refer to a multi-link search which examines only $j \geq 1$ links at each step until no more links can be learned as an *j-link-only search*.

The algorithm proposed in [6], however, did not specify what is the most effective way to combine lookahead search of different number of links. This is the issue we address in this paper. We start by asking the question why some PI models may escape the straightforward multi-link search. The previous example shows that the main reason is the recursive embedding of PI submodels. If a PI submodel $M_1$ is embedded in another PI model $M_2$, $M_1$ will be learned first. After that, if the number of unlearned links in $M_2$ is less than the current number of links to lookahead, $M_2$ will not be learned correctly in the later search steps. In order to learn $M_2$, backtracking to lower number of lookahead links is necessary. Hence the problem translates to a proper arrangement of backtracking during learning.

We propose a multi-link search algorithm (ML) which overcomes the deficiency mentioned above. The learning outcome is represented as DMN. The algorithm focus on learning the chordal structure. Once the chordal graph is obtained, the numerical probability distribution can be estimated from the data.

ML starts with an empty graph. It performs a single link search first. The first stage of the search now ends.

ML then performs a double-link-only search. If some links are learned during the double-link-only search, ML backtracks to perform another single link search. Afterwards, it performs double-link-only search again and backtracks if necessary as before. The combination of double-link-only and single link search will continue until no link is learned in a double-link-only search. We shall refer to this repeated combination of the double-link-only search and the single link search as a *combined-double-link search*. Now the second stage of the search ends.

Next, ML will perform a triple-link-only search. If some links are learned during the search, ML backtracks to repeat the previous two stages. Afterwards, it performs another triple-link-only search and backtracks if necessary as before. We shall refer to this repeated combination of the triple-link-only, double-link-only and single link search as a *combined-triple-link search*. Note that a combined-triple-link search can include several combined-double-link search. Now the third stage of the search ends.

ML continues with a combined-four-link search, followed by a combined-five-link search, etc., until a combined-$k$-link search, where $k > 1$ is a predetermined integer. The pseudo-code of this algorithm is presented below.

**Algorithm ML**
Input: A dataset $D$ over a set $N$ of variables, a
       maximum number $k$ of lookahead links.
Return: The learned graph.
Comment: $lookahead(i)$ is the function for
          an $i$-link-only search.
begin
1        initialize a graph $G = (N, E = \phi)$;
2        for $j := 1$ to $k$ do
3           $i := j$;
4           while $i \leq j$ do
5              $modified :=$ lookahead( $i$ );
6              if $(i > 1)$ AND $(modified = true)$
7              then $i := 1$;    {backtracking}
8              else $i := i + 1$;
9        return $G$ and halts.
end

In algorithm ML, the search stages are indexed by $j$ (line 2) and each iteration of the outer for loop corresponds to one stage. The first iteration has $i = j = 1$ (lines 2 and 3). The single link search $lookahead(1)$ (line 5) will be performed. The test in line 6 will fail and $i$ becomes 2 (line 8). This terminates the while loop as well as the first iteration of the for loop. It corresponds to the first stage of search.

The next iteration of for loop has $i = j = 2$. The double-link-only search $lookahead(2)$ will be performed. If some links have been added, the test in line 6 will succeed and $i$ becomes 1. This causes the execution of another single link search $lookahead(1)$. Afterwards, $i$ becomes 2 and another double-link-only search will be performed. If nothing has been added, $modified$ is false and $i$ becomes 3. This terminates the while loop and the second iteration of the for loop. It corresponds to the second stage of search.

The next iteration of for loop has $i = j = 3$. The triple-link-only search $lookahead(3)$ will be performed. If some links have been added, the test in line 6 will succeed and $i$ becomes 1. This causes the repetition of the previous two stages. This execution of $lookahead(3)$ and repetition of stages 1 and 2 continues until an execution of $lookahead(3)$ returns false. Afterwards, $i$ becomes 4 and the while loop will be terminated. It will also terminate the third iteration of the for loop and end the third stage of search.

The function $lookahead(i)$ performs an *$i$-link-only* search. It consists of multiple *passes* and each pass is composed of multiple *steps*. Each step tests one set of $i$ links. Each pass learns one set of $i$ links after testing all distinct and legal combinations, one at each search step, of $i$ links. This function may be implemented using different scoring metrics. We defer the presentation of our implementation using the cross-entropy scoring metric to section 6.

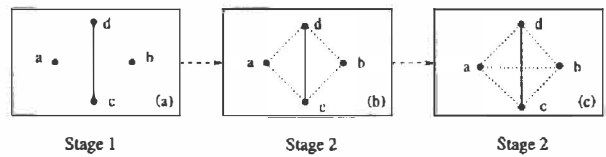

Figure 2: The process of learning the model in Table 1.

Figure 2 shows the execution of ML in learning the PI model in Table 1 with the value of $k$ set as $k = 2$. ML starts with a single link search (The first stage). After all links are examined, one set of links $L_1 = \{(d, c)\}$ is learned. The learned graph is shown in Figure 2 (a). In the second stage, ML performs the double-link-only search first, which learns two sets of links $L_2 = \{(d, a), (a, c)\}$, $L_3 = \{(d, b), (b, c)\}$. These links are contained in the PI submodels over $N_2$ and $N_3$. The corresponding graph is shown in Figure 2 (b). Since some new links are added after the double-link-only search, ML backtracks to perform the single link search again. During this search one set of links $L_4 = \{(a, b)\}$ is added and Figure 2 (c) is obtained. ML continues to perform another double-link-only search but no more links can be learned. The ML halts with a complete graph which is a correct I-map.

## 4  PROPERTY

Can ML learn any PI model correctly? Clearly the answer is no as ML only searchs up to a predetermined number $i$ of lookahead links. A PI submodel that contains more than $i$ colored links may escape ML. Then what is the characteristics of the PI models that can be learned by ML? The following theorem answers this question.

**Theorem 5** *Let $M$ be a PI model such that each embedded PI submodel in $M$ contains no more than $i$ colored links, the algorithm ML with parameter $i$ will return an I-map of $M$.*

Proof:

Let $G_M$ be the minimal colored I-map of $M$. All black links in $G_M$ can be learned by the initial single link search *lookahead*(1) in the first stage. We show that ML will learn colored links in every embedded PI submodel.

When $i = 1$, $M$ contains no embedded PI submodels (no colored links) and the result is trivially true.

When $i = 2$, each embedded PI submodel in $M$ contains only three variables. There are two colored links and one black link among these three variables. The black link will be learned by the initial single link search as mentioned above. The two colored links can be learned by *lookahead*(2) in the second stage.

Now we assume that when $i = k$, if an embedded PI submodel has no more than $k$ unlearned colored links, then these links can be learned by the first $k$ stages. Suppose $i = k+1$. Every PI submodel with no more than $k$ colored links in $M$ can be learned by assumption. For each PI submodel $x$ with $k+1$ colored links in $M$, $x$ either contains one or more embedded PI submodels or contains none.

If $x$ contains at least one embedded PI submodel $y$ of $j \geq 2$ colored links, then we have $j \leq k$ and $y$ must have been learned in the first $k$ stages by assumption. Since the number of remaining colored links in $x$ is $k + 1 - j \leq k - 1$, these links must also have been learned in the first $k$ stages by assumption.

If $x$ contains no embedded PI submodel, then it can be learned by *lookahead* $(k+1)$ at the beginning of stage $k+1$. The theorem is proven. □

Given the parameter $k$ for ML, some PI submodels with more than $k$ colored links may still be learned. Suppose a PI submodel $x$ has more than $k$ colored links and has two other PI submodels $y$ and $z$ embedded in it. If the number of colored links in $y$ or $z$ is no more than $k$, then $y$ and $z$ can be learned by ML. If the number of remaining colored links in $x$ is no more than $k$, then $x$ can also be learned by ML. A formal treatment of such cases will be included in a longer version of this paper.

## 5 COMPLEXITY ANALYSIS

For each pass in an $i$-link-only search, $O(N^{2i})$ sets of $i$ links need to be tested, one set at each step. Therefore each pass contains $O(N^{2i})$ steps. Since each pass adds one set of $i$ links, an $i$-link-only search contains $O(\frac{N^2}{i})$ passes.

Table 2 shows the relation among the index $i$, the number of steps per pass and the number of passes in an $i$-link-only search.

Table 2: The relation among $i$, number of steps per pass and number of passes in an $i$-link-only search.

| $i$ | # of steps/pass | # of passes |
|---|---|---|
| 1 | $O(N^2)$ | $O(N^2)$ |
| 2 | $O(N^{2*2})$ | $O(\frac{N^2}{2})$ |
| 3 | $O(N^{2*3})$ | $O(\frac{N^2}{3})$ |
| . | ... | ... |
| $k-1$ | $O(N^{2*(k-1)})$ | $O(\frac{N^2}{k-1})$ |
| $k$ | $O(N^{2*k})$ | $O(\frac{N^2}{k})$ |

In order to derive the upper bound of the total number of passes in a $k$-link search, we construct a directed graph such that each node in the graph corresponds to one pass during the search and each arrow indicates the chronological order of successive passes. We shall label each node by the number of links to lookahead in the pass. For example, a pass in a single link search will be labeled by 1, and a pass in a double-link-only search will be labeled by 2, etc. A graph so constructed will be a directed chain. For the purpose of a later conversion, nodes with the same label will be drawn at the same level and levels are arranged in the decreasing order of the labels. Figure 3 shows such a graph for the execution of a 3-link-search. The four nodes in the bottom left correspond to the four passes in the first stage during the search. The next three nodes (labeled 2) correspond to the three passes in the first double-link-only search. Since links are learned, they are followed by backtracking to a single link search, shown by the three nodes labeled 1 in the middle bottom of the graph.

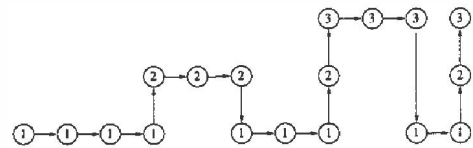

Figure 3: The execution chain of a 3-link search.

Once we obtain such a chain, it can be converted into a set of trees (a forest) as follows. Each node not at the top level will be assigned a parent at the next higher level, and the child and the parent will be connected by an undirected link. The parent of a node is assigned as the first node in the next higher level down the chain. For example, the first node labeled 2 in the chain will be the parent of the first four nodes labeled 1 in the chain. The first three nodes labeled 2 in the chain will have the first node labeled 3 as their parent. After each node not at the top level has been assigned a parent, we remove all arrows from the graph. The resultant graph is shown in Figure 4. Each component of the graph is a tree. This is because each node not at the top level has a unique parent. We shall refer to the graph as an *execution forest*.

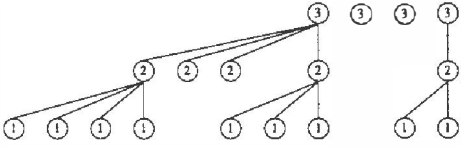

Figure 4: The execution forest of a 3-link search.

We now use the execution forest to analyze the complexity of an i-link-search. For each node at level $i$ ($1 < i \leq k$), some child nodes correspond to learning passes each of which adds a set of $i-1$ links. Other child nodes correspond to non-learning passes that add no links. The number of non-learning passes can not be more than the number of learning passes. The number of learning passes is bounded by $O(\frac{N^2}{i-1})$ according to Table 2. Hence each node at level $i$ ($1 < i \leq k$) has $O(\frac{N^2}{i-1})$ children.

Next, we derive the number of passes at each level. The number of passes at the level $k$ (top level) is $O(\frac{N^2}{k})$. The number of passes at the level $k-1$ is

$$O(\frac{N^2}{k} * \frac{N^2}{k-1}) = O(\frac{N^{2*2}}{k*(k-1)}).$$

The number of passes at the level 2 is

$$O(\frac{N^2}{k} * \frac{N^2}{k-1} * \ldots * \frac{N^2}{2}) = O(\frac{N^{2*(k-1)}}{k*(k-1)*\ldots*2}).$$

Finally, the number of passes at the level 1 is

$$O(\frac{N^2}{k} * \frac{N^2}{k-1} * \ldots * \frac{N^2}{2} * N^2) = O(\frac{N^{2*k}}{k*(k-1)*\ldots*2*1}).$$

Therefore, according to Table 2, the total number of search steps is

$$\begin{aligned}
&O(N^{2*k}) * O(\frac{N^2}{k}) \\
+ &O(N^{2*(k-1)}) * O(\frac{N^{2*2}}{k*(k-1)}) \\
+ &\ldots \\
+ &O(N^{2*2}) * O(\frac{N^{2*(k-1)}}{k*(k-1)*\ldots*2}) \\
+ &O(N^2) * O(\frac{N^{2*k}}{k*(k-1)*\ldots*2*1}) \\
= &O((N^{2*(k+1)}) * (\frac{1}{k} + \frac{1}{k*(k-1)} + \ldots + \frac{1}{k*(k-1)*\ldots*2*1})).
\end{aligned}$$

Since the factor $(\frac{1}{k} + \frac{1}{k*(k-1)} + \ldots + \frac{1}{k*(k-1)*\ldots*2*1})$ is upper-bounded by 1, the total number of search steps in a $k$-link-search is $O(N^{2*(k+1)})$.

In order to complete the complexity analysis, we need to take into account of the complexity of each search step, which is dependent on the choice of scoring metric used in $lookahead(i)$. Our implementation, to be detailed in the next section, is based on the algorithm in [8]. The complexity of one search step is

$$O(n + \eta(\eta \log \eta + 2^\eta)),$$

where $n$ is the number of cases in the dataset and $\eta$ is the maximum size of cliques. Hence the overall complexity of the algorithm is

$$O(N^{2(k+1)}(n + \eta(\eta \log \eta + 2^\eta)).$$

Compared with the complexity of a straightforward multi-link search algorithm [8]

$$O(k\, N^{2k}(n + \eta(\eta \log \eta + 2^\eta)),$$

the complexity of a $k$-link-search using ML is higher but not much higher. The benefit of the slightly increased complexity is the capability of learning recursively embedded PI models.

## 6  IMPLEMENTATION AND EXPERIMENTAL RESULTS

Given the algorithm ML, the only missing detail in implementation is the function $lookahead(i)$. Our implementation of this function is based on the algorithm in [8]. Instead of testing the conditional independence directly, a test of whether new links decrease the Kullback-Leibler cross entropy is performed. This is justified the following shown in [8]. (1) Minimizing the K-L cross entropy between a dataset $D$ and a DMN obtained from $D$ is equivalent to minimizing the entropy of the DMN. (2) A learning process starting with an empty DMN structure and driven by the minimization of the above K-L cross entropy is paralleled by the process of removing false independence (missing links relative to some minimal I-map) in the intermediate DMNs.

The pseudo code of the $lookahead(i)$ function is shown below. A threshold $\delta$ is used to differentiate between a strong dependence and a weak one (may be due to noise). A greedy search can thus be applied (line 4 through 9) to avoid adding unnecessary links and links due to weak dependence [8]. The condition that $L$ is implied by a single clique $C$ means that all links in $L$ are contained in the subgraph induced by $C$. This requirement helps to reduce the search space.

```
Function BOOL lookahead( int i );
Input: i is the number of lookahead links.
Comment: δh is a threshold.
begin
1   modified := false;
2   repeat
3     initialize the entropy decrement dh' := 0;
4     for each set L of links (|L| = i, L ∩ E = φ), do
5       if G* = (N, E ∪ L) is chordal and L is implied by a
6       clique, then compute the entropy decrement dh*;
7       if dh* > dh', then dh' := dh*, G' := G*;
8     if dh' > δh, then G := G', done := false,
9        modified := true;
10    else done := true;
11  until done = true;
12  return modified;
end
```

The following demonstrates our implementation with two datasets. Our primary emphasis is the capability of learning correctly PDMs with recursively embedded PI submodels. First, a dataset of 1000 cases was generated from the PDM shown in Table 1. The successful run used $k = 2$, $\delta h = 0.001$. The learning process is the same as Figure 2. It is summarized in Table 3.

Table 3: Summary of learning the PDM in Table 1

| $i-link-$ only search | learned link set | # graphs tested | cross entropy decrement |
|---|---|---|---|
| 1 | $\{(d,c)\}$ | 6 | 0.0033 |
| 2 | $\{(d,a),(a,c)\}$ | 26 | 0.0139 |
| 2 | $\{(d,b),(b,c)\}$ | 29 | 0.0022 |
| 1 | $\{(a,b)\}$ | 30 | 0.0389 |

Next, we use a PDM from [6] described below:

Three balls are drawn each from a different urn. Urn 1 has 20% white balls and the rest of the balls black. Urn 2 and urn 3 have 60% and 50% of white balls, respectively. A music box plays if all three balls are white or exactly one is white. A dog barks if two random lights are both on or both off. John complains if it's too quiet (neither the box plays nor the dog barks) or too noisy (both the box plays and the dog barks).

The model is specified as a Bayesian network shown in Figure 5. Its colored I-map is shown in Figure 6.

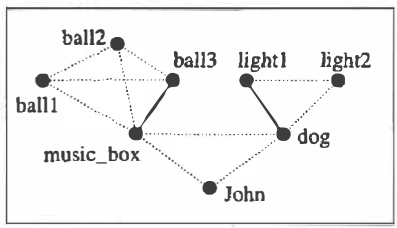

Figure 6: Colored I-map of the music-box example.

The PDM contains five embedded PI submodels over

$$N_1 = \{ball1, ball3, music\_box\},$$
$$N_2 = \{ball2, ball3, music\_box\},$$
$$N_3 = \{ball1, ball2, ball3, music\_box\},$$
$$N_4 = \{light1, light2, dog\},$$
$$N_5 = \{music\_box, dog, John\}.$$

Note that the first two PI submodels are recursively embedded in the third PI submodel.

We generated a dataset of 2000 cases from the music-box-dog-John domain. Using $k = 3$ and $\delta h = 0.004$, the algorithm learned the I-map successfully. The learning process is shown in Figure 7.

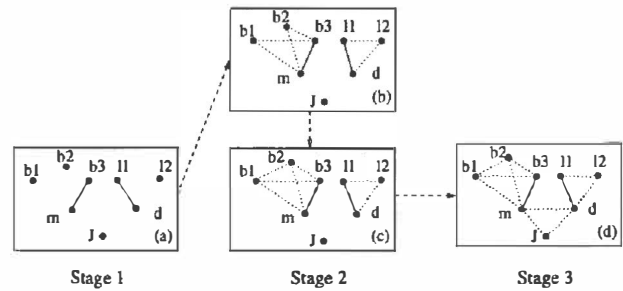

Figure 7: The process of learning the music-box model.

The algorithm started by performing the single link search. In the first pass, one link was learned:

$$L_1 = \{(light1, dog)\}$$

It took 28 steps (28 candidate graphs tested). In the second pass, after 27 steps, another link was learned:

$$L_2 = \{(ball3, music\_box)\}.$$

Note that a standard single-link search learning algorithm will halt and returns this graph which is not an I-map of the domain. Since nothing was learned in the third pass, a 2-link-only search was performed next. After 884 steps, three sets of links were learned in the following order:

$$L_3 = \{(light1, light2), (light2, dog)\},$$
$$L_4 = \{(ball2, ball3), (ball2, music\_box)\},$$
$$L_5 = \{(ball1, ball3), (ball1, music\_box)\}.$$

Then the algorithm backtracked to perform a single link search with one link learned:

$$L_6 = \{(ball1, ball2)\}.$$

During the next single link search and the following 2-link-only search, no link was added. Hence a 3-link-only search was performed, which learned the links:

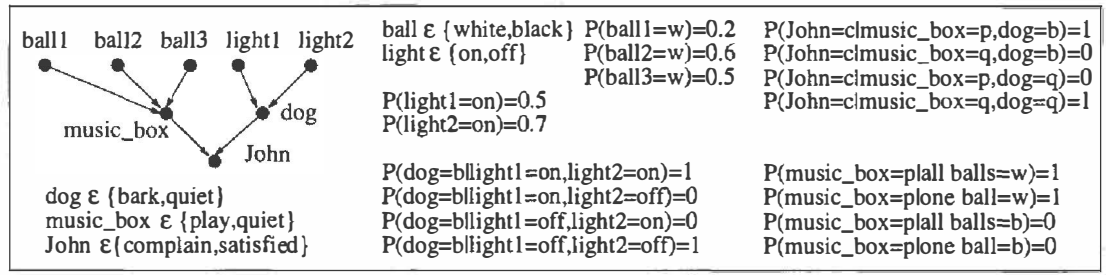

Figure 5: The specification of the music-box model.

$L_7 = \{(music\_box, dog), (dog, John), (John, music\_box)\}$.

The backtracking occurred afterwards, but no more links was learned. Finally, the algorithm halted and returned the correct I-map. A total of 3583 candidate graphs were tested. A summary of the experiment is shown in Table 4.

Table 4: Summary of learning result

| $i - link - only$ search | learned link set | # graphs tested | cross entropy decrement |
|---|---|---|---|
| 1 | $L_1$ | 28 | 0.0822 |
| 1 | $L_2$ | 55 | 0.0069 |
| 2 | $L_3$ | 432 | 0.6109 |
| 2 | $L_4$ | 708 | 0.1922 |
| 2 | $L_5$ | 939 | 0.0146 |
| 1 | $L_6$ | 1149 | 0.4802 |
| 3 | $L_7$ | 2327 | 0.6895 |

## 7 CONCLUSION

PI models escape the detection of many algorithms for learning belief networks that rely on a single link search to detect local dependency. They form a class of difficult PDMs for automated learning. PI models do exist in practice with *parity* problems and *modulus addition* problems as special examples [6]. Earlier work by Xiang et al. [8] proposed a straightforward multi-link search algorithm to learn PI models. In this work, we show that when a PDM contains recursively embedded PI submodels, it may escape the straightforward multi-link search algorithm. We have presented an improved algorithm that learns a DMN as an I-map of a domain with recursively embedded PI submodels. We have shown that the algorithm will uncover all embedded PI submodels as long as the size of the submodel is within a predetermined bound. The performance of the algorithm is demonstrated with experiments.

We have also analyzed the complexity of the improved algorithm. The result shows that the improved learning capability of the new algorithm only cause slight increase in the complexity compared with the straightforward multi-link search algorithm.

We believe that no search steps in the improved algorithm may be deleted without jeopardizing the above learning capability. We are currently working to formally establish this result.

## Acknowledgements

This work is supported by grant OGP0155425 from the Natural Sciences and Engineering Research Council and grant from the Institute for Robotics and Intelligent Systems in the Networks of Centres of Excellence Program of Canada.